\documentclass{article}

\usepackage{microtype}
\usepackage{graphicx}
\usepackage{subfigure}
\usepackage{multirow}
\usepackage{booktabs} 
\usepackage{hyperref}
\usepackage{amsmath,amssymb,amsthm}
\usepackage{verbatim,float,url,enumerate}
\usepackage{graphicx,subfigure,psfrag}
\usepackage{natbib,xcolor}
\usepackage{microtype}
\usepackage{graphics}
\usepackage{tikz}
\usetikzlibrary{bayesnet}
\usetikzlibrary{arrows}
\usetikzlibrary{backgrounds,positioning}
\usetikzlibrary{decorations.pathreplacing}
\usepackage{caption}
\usepackage{wrapfig,lipsum,booktabs}
\usepackage{bm}
\usepackage{algorithm}
\usepackage{algorithmic}
\usepackage{enumitem}

\newtheorem{assumption}{Assumption}

\usepackage{hyperref}

\usepackage[accepted]{icml2020}

\icmltitlerunning{Covariate Distribution Aware Meta-learning}

\begin{document}

\twocolumn[
\icmltitle{Covariate Distribution Aware Meta-learning}



\icmlsetsymbol{equal}{*}

\begin{icmlauthorlist}
\icmlauthor{Amrith Setlur}{equal,to}
\icmlauthor{Saket Dingliwal}{equal,to}
\icmlauthor{Barnabas Poczos}{to}
\end{icmlauthorlist}
\icmlaffiliation{to}{Carnegie Mellon University}
\icmlcorrespondingauthor{Amrith Setlur}{asetlur@cs.cmu.edu}
\icmlkeywords{Meta-learning, Few-shot learning}
\vskip 0.3in
]

\printAffiliationsAndNotice{\icmlEqualContribution}






\newcommand*{\thead}[1]{%
\multicolumn{1}{c}{\bfseries\begin{tabular}{@{}c@{}}#1\end{tabular}}}

\def\V#1{\mathbf{#1}}
\def\C#1{\mathcal{#1}}
\def\R#1{\mathrm{#1}}
\def\B#1{\mathrm{#1}}

\definecolor{Red}{rgb}{1,0,0}
\definecolor{Green}{rgb}{0,0.7,0}
\definecolor{Blue}{rgb}{0,0,1}
\definecolor{Red}{rgb}{0.6,0,0}
\definecolor{Orange}{rgb}{1,0.5,0}

\newcommand{\am}[1]{\textcolor{Blue}{[#1 --Amrith]}}
\newcommand{\sd}[1]{\textcolor{Green}{[#1 --Saket]}}
\begin{abstract}
    Meta-learning has proven to be successful for few-shot learning across the regression, classification, and reinforcement learning paradigms. 
    Recent approaches have adopted Bayesian interpretations to improve gradient-based meta-learners by quantifying the uncertainty of the post-adaptation estimates. Most of these works almost completely ignore the latent relationship between the covariate distribution $(p(x))$ of a task and the corresponding conditional distribution $p(y|x)$. In this paper, we identify the need to explicitly model the meta-distribution over the task covariates in a hierarchical Bayesian framework. We begin by introducing a graphical model that leverages the samples from the marginal $p(x)$ to better infer the posterior over the optimal parameters of the conditional distribution $(p(y|x))$ for each task. Based on this model we propose a computationally feasible meta-learning algorithm by introducing meaningful relaxations in our final objective. We demonstrate the gains of our algorithm over initialization based meta-learning baselines on popular classification benchmarks. Finally, to understand the potential benefit of modeling task covariates we further evaluate our method on a synthetic regression dataset.  
\end{abstract}

\section{Introduction}
\label{sec:introduction}

\begin{figure*}[htp]
  \centering
  \subfigure[\label{fig:reg}Samples of tasks drawn from a meta-distribution categorized based on its corresponding hypothesis class. The crosses indicate the labeled sampled points and the dashed lines represent the true hypothesis for different tasks. Clearly, the choice of the hypothesis class and thereby the parameters for $y|x$ depend on the covariate distribution of the task.]{\includegraphics[width=0.61\textwidth]{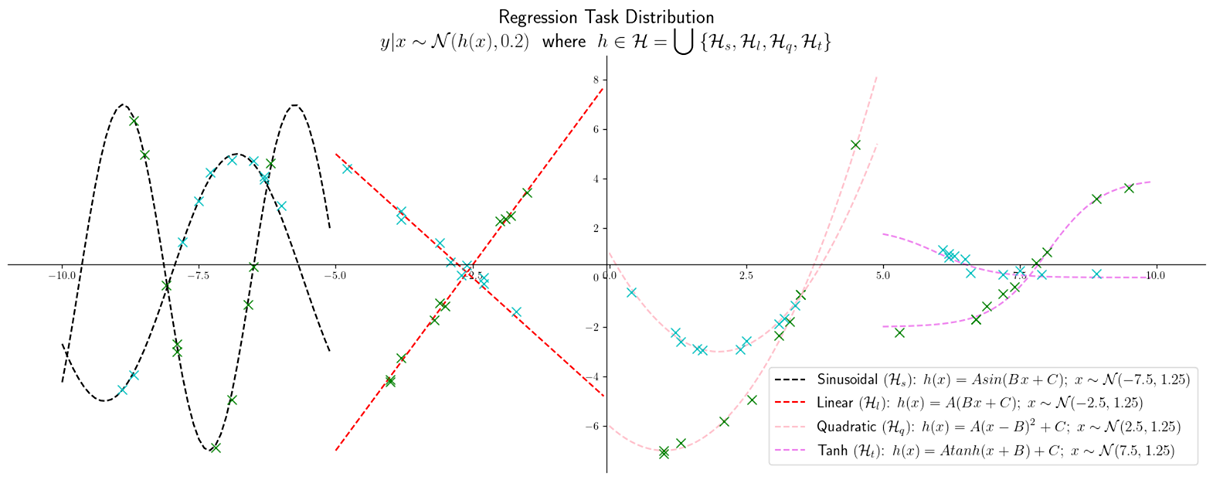}}\quad
  \subfigure[\label{fig:class}For image classification, since $p(x)$ in  Test Task 1 and Train Task 2 would be overlapping (all are animals), our hypothesis is that the discriminative features for these will also be related.]{\includegraphics[width=0.35\textwidth]{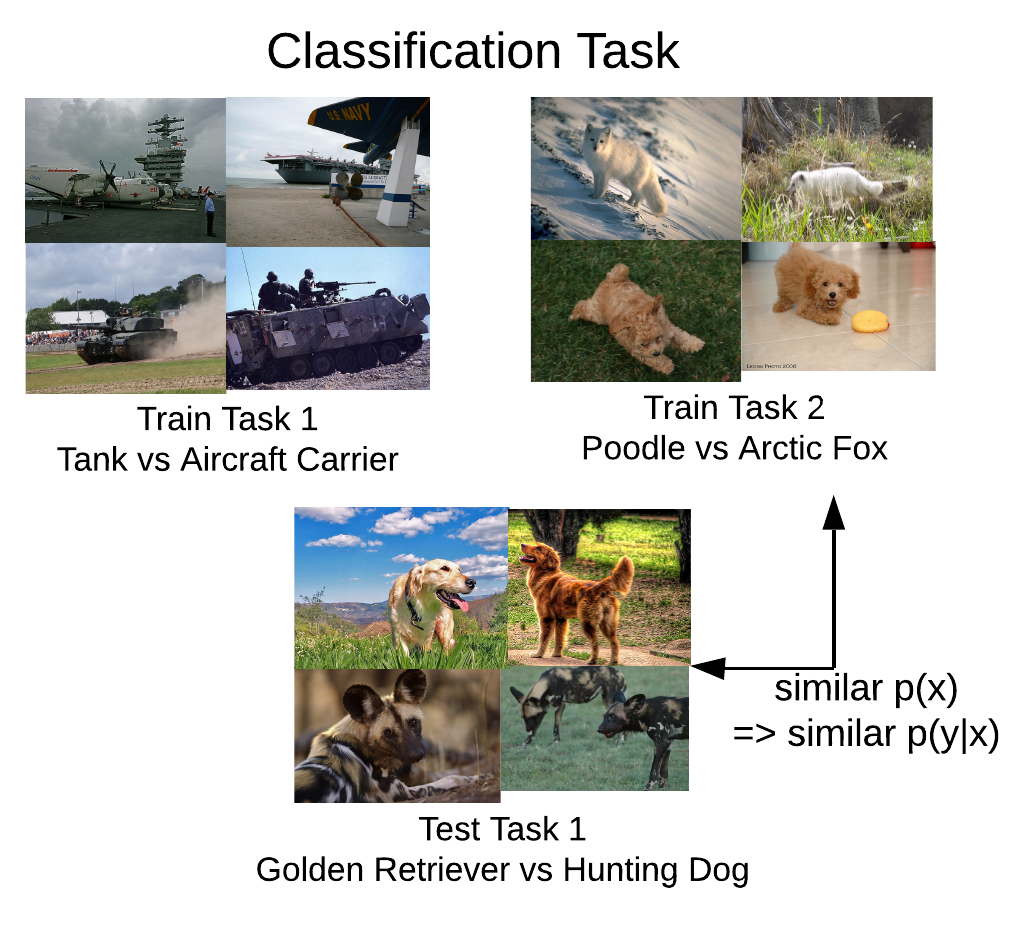}}
  \vspace{-1em}
  \caption{Regression (a) and classification (b) examples to motivate the potential benefits of meta-distributions over covariates.}
  \vspace{-1em}
\end{figure*}


Learning quickly or with very few samples has been a long-term goal of the machine learning community. The field of meta-learning has recently made significant strides towards achieving that goal. Meta-learning \cite{nichol2018first, ravi2016optimization, finn2017model} comprises of a set of algorithms designed to exploit prior experiences from multiple tasks (drawn from a task distribution) for improving sample-efficiency on a new but related task from the same distribution. Given the increasing cost of getting annotated samples on an ever-increasing variety of related tasks, the practical scope of these algorithms is immense.

Most meta-learning methods can be classified into two broad categories (i) \textit{gradient-based} \cite{ravi2018amortized, denevi2019learning, finn2017model} approaches that meta-learn parameters of optimization algorithms (like initialization and learning rate) in a way that the meta-learner (optimizer) is amenable to quickly adapt on a new task by performing gradient descent on a very small number of labeled samples, and (ii) \textit{amortized-inference} \cite{snell2017prototypical, lee2019meta, bertinetto2018meta} based approaches that directly infer the optimal parameters of a new task without performing any gradient based optimization. 
In this work, although we focus on improving gradient-based methods, we believe that our core idea can be adapted to the latter as well. 
Recent works \cite{finn2018probabilistic}, \cite{ravi2018amortized}, \cite{kim2018bayesian} have used a Bayesian framework to learn a suitable prior over the network parameters by leveraging the inherent structure of the task distribution. 
By viewing the parameters of a meta-learner through a Bayesian lens we can use the  posterior \cite{gal2016dropout} to estimate the uncertainty of the adapted parameters for each task \cite{ravi2018amortized}.

In a Bayesian meta-learner, 
the posterior over the adapted network parameters for a new task is typically inferred using a few samples from the task along with a meta-learned prior. In this work, we hypothesize that the covariate distribution of a task can also influence the posterior over the adapted network parameters. To the best of our knowledge none of the existing meta-learning algorithms like \citet{bertinetto2018meta, rajeswaran2019meta, ravi2018amortized, finn2018probabilistic} explicitly utilize the information present in the covariates to improve the estimate of the adapted parameters. 
We do this by modeling the latent factors of the covariate distribution.
We define a prior not only on the network parameters (which determine the conditional $p(y|x)$) but also on the covariate distribution $p(x)$. 
Our meta-learning objective involves maximizing the joint likelihood $p(x, y)$ as opposed to just $p(y|x)$ 
which leads to meta-parameters sharing information about the covariates across tasks, in addition to the optimal network parameters. 
This way the latent factors of the covariate distribution $p(x)$ of a new task can be quickly inferred from very few covariates. Finally, the inferred latent covariate factors are used to infer the posterior over the adapted network parameters.



The main contributions of our work are as follows: (1) we identify and motivate the need to model the latent structure present in the covariate distributions $(p(x))$ for a sequence of tasks (2) to the best of our knowledge we are the first to propose a Bayesian framework which exploits this latent information to better infer the posterior over the adapted network parameters  (that define $p(y|x)$) (3) we propose a gradient based model-agnostic meta-learning algorithm that is an instantiation of our probabilistic theory and demonstrate its benefits on popular classification datasets and synthetic regression datasets.  



\section{Motivation}
\label{sec:motivation}

Recent work in Semi-Supervised Learning(SSL) for classification \cite{laine2016temporal, miyato2018virtual, wang2019enaet, sohn2020fixmatch} and regression \cite{jean2018semi} have established the significance of unlabelled samples (and hence the covariate distribution) in the respective tasks. In fact, in certain cases, SSL approaches the performance of purely supervised learning, even when a substantial portion of the labels in a given dataset has been discarded \cite{oliver2018realistic}. Particularly, for high-dimensional data like images with limited labelled samples, the role of $p(x)$ in finding optimal network parameters is immense.

Taking inspiration from SSL, we identified that performance of a meta-learner can benefit from learning the meta-distribution over the covariates of the training tasks. Given this meta-distribution and samples from the test task, we can identify the manifold from which $p(x)$ of the test task is sampled and hence choose a better informed prior for $p(y|x)$. For simplicity, we first motivate it with a synthetic example as in Figure \ref{fig:reg}. Assume that for all the tasks, $p(y|x) = \C{N}(h(x), 0.2)$ where the optimal hypothesis $h \in \C{H}$$=$$\bigcup \; \{\C{H}_s, \C{H}_l, \C{H}_q, \C{H}_t\}$ that can be classified into four hypothesis classes: \textit{sinusoidal} ($\C{H}_s$), \textit{linear} ($\C{H}_l$), \textit{quadratic} ($\C{H}_q$) and \textit{tanh} ($\C{H}_t$). Each of the hypothesis classes have multiple hypothesis with different parameters in them $(h_1 = 2 sin(3x + 2), h_2 = -2 sin (4x + 3) \text{ then } h_1,h_2 \in \C(H)_s )$. Further, assume that the input distribution $p(x)$ is vastly different for each of the four hypothesis classes. For example let us say $x \sim \C{N}(-7.5,1.25)$ whenever $h \in \C{H}_s$ and $x \sim \C{N}(2.5,1.25)$ whenever $h \in \C{H}_q$. Now, given a target task if we observe covariates in the range $[-10, -5]$, it is highly likely that true hypothesis for this task lies in $\C{H}_s$. Therefore, modelling the meta-distribution over covariates can help to choose a prior for $p(y|x)$ such that it have a higher measure for $\C{H}_s$. We empirically verify this claim in the Experiment section.


The intuition used in the above mentioned setting extends beyond the example. In few-shot image classification, images for different tasks can lie on different manifolds \cite{saul2003think}. For example, as shown in Figure \ref{fig:class}, miniImageNet images from classes like arctic-fox, miniature-poodle intuitively lie in a very different manifold ($p(x)$) than classes like tanks, aircraft-carrier etc. Intuitively, the discriminative features for classifying animals will also be very different from features for heavy vehicles. Now, modelling the meta-distribution over covariates for different training tasks can help the meta-learner to understand the manifold structure of the image space. Then, for a test task, where we need to classify hunting-dog vs golden-retriever, the meta-learner can identify the corresponding manifold of the covariates(animals here) and set a prior for network parameters that have high measure for discriminative features specific to animals. Such a task-specific prior on $p(y|x)$ can boost the performance of the learner to adapt quickly using very few examples.

\section{Related Work}
\label{sec:related-work}

Our methodology is complementary to most existing works in the probabilistic meta-learning literature. We borrow the basic hierarchical Bayes framework from \citet{ravi2018amortized, finn2018probabilistic} and extend it to model Bayesian variables that generate the covariate distribution for a task. This enables our method to be model-agnostic while having the ability to benefit from the latent relationship between the task covariates and the optimal parameters.
In the non-Bayesian setting, the \textsc{m-maml} algorithm proposed by \citet{vuorio2019multimodal} is mildly similar to our approach in the sense that they learn task specific initializations instead of a single one as originally introduced by \citet{finn2017model}. \textsc{m-maml} uses the labeled samples to choose an initialization for a given task and hence one can view the covariate distribution as being used indirectly. 
While they try to identify the mode of a task to be able to choose a better initialization, we explicitly model the meta-distribution over covariates of different tasks and hence mutual information between the covariates and network parameters is captured in a more general way.   
Our approach is more direct since it \textit{first} infers the posterior over the latent factors of the covariate distribution via a maximum likelihood objective and then uses the inferred posterior to improve the adaptation of network parameters. Additionally, thanks to our Bayesian framework, we are capable of modeling the uncertainty of the adaptation which can prove to be be critical in the few shot scenario. 

\section{Methodology}
\label{sec:methodology}
We begin by introducing some notations for the meta-learning setup used in the rest of the paper followed by the proposed probabilistic framework which explicitly exploits (i) the structure of the covariate distributions across tasks (ii) the relation between the covariate distribution and optimal hypothesis for a given task. We then derive the Maximum Likelihood Estimation (\textsc{mle}) objectives for the observed variables in our model.
Finally, we discuss a specific meta-learning algorithm that can efficiently optimize the proposed objective. We do this via an instantiation of the generic approach obtained by making certain simplifying assumptions in the original framework.

\textbf{Notations} We are given a sequence of $n$ tasks $\{\C{T}_i\}_{i=1}^{n}$ with each task $\C{T}_i$ having $m$ labeled samples given by the dataset $\C{D}_i = \{\V{x}_j^{(i)}, \V{y}_j^{(i)}\}_{j=1}^{m}$ where $\V{x}_j^{(i)} \in \C{X} \subset \B{R}^{k}$ and $\V{y}_j^{(i)} \in \C{Y}\,\, \subset \B{R}$. Following the definitions introduced by \citet{finn2018probabilistic} we split the dataset  $\C{D}_i := \{\C{D}_i^{S}, \C{D}_i^{Q}\}$ into support ($\C{D}^{S}_i$) and query $(\C{D}^{Q}_i)$ sets respectively with $|\C{D}_i^S|=m', |\C{D}_i^Q|=m-m'$. Each sample in $\C{D}_i$ is drawn from the joint distribution $p_i(x, y)$ over $\C{X} \times \C{Y}$ with the marginals given by $p_i(x)$ and $p_i(y)$. 

\begin{figure}[!htbp]
    \centering
    \includegraphics[width=0.6\linewidth]{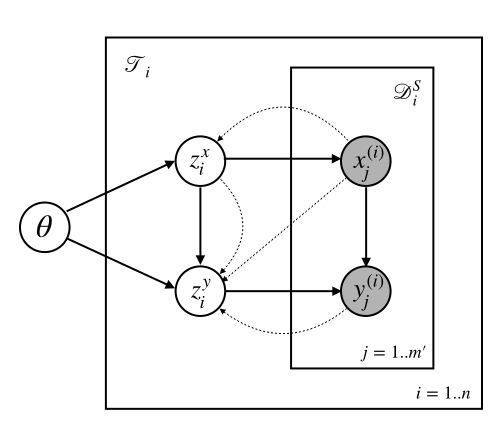}
    \caption{A graphical model representing our hierarchical Bayes framework with task-agnostic meta-parameters $\theta$ and task-specific latent variables $z^x_i$, $z^y_i$ which influence the marginal $p_i(x)$ and conditional $p_i(y|x)$ distributions respectively. The dotted lines denote the variational approximations introduced over the posteriors. 
    The objective remains to learn variational parameters that suitably maximize the likelihood of the observed variables (shaded).
    }
    \label{fig:gm_CPMAML}
\end{figure}

The probabilistic model we consider in our work is summarized in Fig.~\ref{fig:gm_CPMAML}. Without making any assumptions on the nature of $p_i(x, y)$, we assume the existence of meta-parameters $\theta$ that govern the common structure shared across the set of joint distributions $\{p_{i}(x,y)\}_{i=1}^{n}$.
Within each task the generative model for the input $x^{(i)}$ involves a random variable $z^{x}_{i}$ which we shall refer to as the latent factors of the covariate distribution.
Also, each task has an additional latent variable $z^{y}_{i}$ which plays a role in the generative model for the response variable $y^{(i)}$ given the input $x^{(i)}$.
In most settings, a naive assumption of independence is made over the latent factors $z^{x}_{i}$ and $z^{y}_{i}$. Because of this assumption and the fact that we are only interested in $p(y|x)$ of a target task, the latent factors $z^{x}_{i}$ are completely ignored in the existing frameworks. On the other hand, motivated by the reasoning in Motivation section, we refrain from making such an assumption and instead exploit the information present in the covariates $\{\V{x}_j^{(i)}\}_{j=1}^{m}$ to better infer the posterior over the latent variable $z^{y}_i$ which influences the conditional $p_i(y|x)$.


\subsection{Formal Derivations}
\label{subsec:formal_derivations}
In this section, we derive a lower bound for the likelihood of the observed data $\{\C{D}_i^{S}\}_{i=1}^{n}$ using hierarchical variational inference. This gives us the meta-learning objective that can be optimized using standard gradient-based approaches.
\begin{align}
    \label{eq:deriv-1}
    \log &\,\, p(\{ \C{D}_i^{S} \}_{i=1}^n) = \log \int_\theta p(\{\C{D}_i^{S}\}_{i=1}^n | \theta) p(\theta) \; d\theta   \\
    & \geq \R{E}_{q(\theta; \beta)} \left[ \sum_{i=1}^{n}\log p(D_i^{S} | \theta) \right] - \textsc{kl}(q(\theta; \beta)  || p(\theta)) \nonumber
\end{align}
In the above equation, the distribution $q(\theta; \beta)$ is a variational approximation (with parameters $\beta$) for the true posterior over the meta-parameter $\theta$. For the derivations henceforth we shall drop the notations $i$ and $S$ when understood from context. The log-likelihood of the dataset $\C{D}_i$ given by $p(\C{D}_i|\theta)$, can be written as an integral over the factors $p(\C{D}_i|z^x_{i}, z^{y}_{i})$, $p(z^{y}_{i}|z^{x}_{i}, \theta)$ and  $p(z^{x}_{i} | \theta)$.
\begin{align}
   p(\C{D}_i|\theta) \hspace{-0.1em} = \hspace{-0.1em} \int_{z^x_i} \hspace{-0.3em}\int_{z^y_i} p(\C{D}|z^x_i, z^y_i) p(z^y_i|z^x_i, \theta) p(z^x_i|\theta)\;\;  dz^y_i dz^x_i \nonumber
\end{align}
To lower bound the log of the above objective we introduce two variational approximations (i) $q(z^x_i; \kappa_i)$ with parameters $\kappa_i$ for the true posterior $p(z^x_i | \{\V{x}_j^{(i)}\}_{j=1}^{m'})$ and (ii) $q(z^y_i; \lambda_i)$  with parameters $\lambda_i$ for the true posterior $p(z^y_i | z^x_i, \C{D}_i^{S})$.
\begin{align}
     \log p(\C{D}_i|\theta) & \hspace{-0.2em} \geq  \hspace{-0.2em} \R{E}_{q(z^x_i;\kappa_i)} \hspace{-0.4em} \left[ \log \int p(\C{D}_i | z^x_i, z^y_i) p(z^y_i | z^x_i, \theta)  dz^y_i \right] \nonumber \\  
     & - \textsc{kl}(q(z^x_i;\kappa_i) || p(z^x_i|\theta))
     \label{eq:deriv-elbo-zx}
\end{align}
Since $z^x_i$ is the latent factor in the generative model for $x^{(i)}$ and $z^{y}_{i}$ is the corresponding latent variable for $y^{(i)}$, from  graphical model in Fig. \ref{fig:gm_CPMAML}, we arrive at the following independence: (1) $\V{y}_{j}^{(i)} \perp z^x_i | z^y_i,  \V{x}_{j}^{(i)}$ and (2) $\V{x}_{j}^{(i)} \perp z^y_i| z^x_i$. Based on this, we break $p(\C{D}_i | z^x_i, z^y_i)$ as in Eq. \ref{eq:break_pd} 
\begin{align}
\label{eq:break_pd}
   p(\C{D}_i | z^x_i, z^y_i) = \prod_{j=1}^{m^{'}} p(\V{y}_{j}^{(i)} |  \V{x}_{j}^{(i)}, z^{y}_{i}) p(\V{x}_{j}^{(i)} | z^x_i)
\end{align}
We finally arrive at the following Evidence Lower Bound (\textsc{elbo}) for $\log p(\C{D}_i | \theta)$ which we shall refer to as $\C{L}_{\C{D}_i}(\kappa_i, \lambda_i, \theta)$. 
\begin{align}
\label{eq:deriv-elbo-zy}
   & \C{L}_{\C{D}_i} (\kappa_i, \lambda_i, \theta) = \R{E}_{q(z^x_i;\kappa_i)}\left[  \sum_{j=1}^{m'} \log p(\V{x}_j^{(i)} | z^x_i) \right]  \\
    - & \textsc{kl}(q(z^x_i;\kappa_i)  ||  p(z^x_i|\theta)) +  \R{E}_{q(z^x_i;\kappa_i)} \;  \C{L}_{\C{D}_i}'(z^x_i, \lambda_i, \theta) \nonumber \\
    \label{eq:deriv-elbo-zy-2}
    & \C{L}_{\C{D}_i}'(z^x_i, \lambda_i, \theta) =   \R{E}_{q(z^y_i;\lambda_i)} 
    \left[ \sum_{j=1}^{m'} \log p(\V{y}_j^{(i)} | \V{x}_j^{(i)}, z^y_i) \right]  \nonumber \\  
     & \quad\quad\quad\quad\quad\quad-  \textsc{kl}(q(z^y_i;\lambda_i)||p(z^y_i|z^x_i, \theta))  
\end{align}

Therefore, the \textsc{elbo} on the likelihood of the dataset for $i^{th}$ task is a function of the task-specific variational parameters $\kappa_i, \lambda_i$ and the meta-parameter $\theta$ (which itself is sampled using variational distribution with task-agnostic parameter $\beta$). As per our formulation, these variational parameters that approximate the true posteriors for each task will be distinct. Hence, $\kappa_i, \lambda_i$ need to be adapted for each task individually. 
Given $ \C{L}_{\C{D}_i} (\kappa_i, \lambda_i, \theta)$ we can re-write the overall lower bound by substituting back in Eq.~\ref{eq:deriv-1} as:
\begin{align}
\label{eq:final-elbo}
& \quad\log \,\, p(\{ \C{D}_i^{(S)} \}_{i=1}^n) \geq \Gamma(\kappa_1,\dots\kappa_n,\lambda_1,\dots\lambda_n,\beta)   \\
& = \R{E}_{q(\theta; \beta)} \left[ \sum_{i=1}^{n} \C{L}_{\C{D}_i}(\kappa_i, \lambda_i, \theta)  \right] - \textsc{kl}(q(\theta; \beta)  || p(\theta)) \nonumber
\end{align}

\subsection{Algorithm}
\label{subsec:algorithm}

The primary aim of any meta-learning algorithm is to optimize for the meta-parameter $\theta$ given the sequence of tasks and the corresponding datasets $\{\C{D}_i\}_{i=1}^{n}$. This is generally a two step process where step-I involves identifying the optimal task-specific parameters using $\theta$ and the support set $\C{D}_i^{S}$. In step-II, based on the task-specific adapted parameters from step-I the meta-parameter $\theta$ is optimized over the query set $\C{D}_i^{Q}$. Within our framework, since both the meta-parameter $\theta$ and the task-specific latent parameters $z^x_i, z^y_i$ are Bayesian random variables with their variational parameters given by $\beta, \lambda_i, \kappa_i$ respectively, we instead define an algorithm to optimize the \textsc{elbo} in Eq.~\ref{eq:final-elbo} to get optimal $\beta^{*}$ using a similar two step process. Note that we use Amortized Variational Inference (AVI) as done in \cite{ravi2018amortized} to arrive at the objective defined in Eq.~\ref{eq:inner-loop}. 
\begin{align}
    \label{eq:outer-loop}
    \beta^* &= \arg\min\limits_{\beta} \;  -\Gamma(\kappa_1,\dots\kappa_n,\lambda_1,\dots\lambda_n,\beta) \\
    &= \arg\min\limits_{\beta} \;  -\R{E}_{q(\theta; \beta)} \sum\limits_{i=1}^n  \C{L}_{\C{D}_i^{Q}}(\kappa_i^*, \lambda_i^*, \theta) \nonumber \\
    & \quad \quad \quad \quad \quad \quad \quad \quad  + \textsc{kl}(q(\theta; \beta)  || p(\theta)) \nonumber  \\
    \label{eq:inner-loop}
    &\kappa_i^* , \lambda_i^* = \arg\min\limits_{\kappa_i, \lambda_i} \;  \C{L}_{\C{D}_i^{S}}(\kappa_i, \lambda_i, \theta)
\end{align}
Note that the theoretical framework developed above for modelling the covariate distribution is very general. Based on choice of variational and the prior distributions, we can instantiate different meta-learning algorithms from our probabilistic theory. To particularly illustrate one such meta-learner, we make some reasonable choices for these distributions following \cite{ravi2018amortized}. For this meta-learner, we show that the optimization of the defined likelihood is computationally feasible and the algorithm is intuitively simple. 
\begin{assumption}
\label{assm:theta}
For making the likelihood objective simple and computationally feasible, we first assume the variational approximation $q(\theta; \beta)$ follows a $\delta-$distribution given by $\delta_{\beta}$ and the prior $p(\theta)$ is given by $\C{N}(\V{0}, \V{I})$. 
\end{assumption}
\begin{assumption}
\label{assm:z_x}
The latent parameters $z^x_i$ for any task with dataset $\C{D}_i^{S}$ are assumed to be normally distribution with mean, diagonal co-variance matrix $\kappa_i^*= (\gamma_\mu(\C{D}_i^{S}; \beta), \gamma_\sigma^2(\C{D}_i^{S}; \beta))$. Note that we make this assumption because we can choose the functions $\gamma_\mu(\cdot), \gamma_\sigma^2(\cdot)$ to be neural networks with parameters $\beta$ giving immense representational power to approximate any posterior distribution \cite{kingma2013auto}. For simplicity, again we can assume the prior on these latent variables ($p(z^x_i|\theta)$) to be $\C{N}(\V{0}, \V{I})$. 
\end{assumption}
\begin{assumption}
\label{assm:z_y}
We choose the distribution $q(z^y_i; \lambda_i)$ to be given by a $\delta-$distribution: $\delta_{\lambda_i}$. Intuitively, $q(z^y_i; \lambda_i)$ approximates the true posterior for latent parameters $z^y_i$ after observing $\{ \V{x}_{j}^{(i)},\V{y}_{j}^{(i)}\}_{j=1}^{m^{'}}$. Now if we assume $p(\V{y}_{j}^{(i)} |  \V{x}_{j}^{(i)}, z^{y}_{i}) = \delta( f(\V{x}_{j}^{(i)}; z^{y}_{i} ))$ where $f(\cdot)$ is a neural network with parameters $z^{y}_{i}$, this variational approximation is reasonable. 
\end{assumption}


Using Assm.~\ref{assm:theta}, the Eq.~\ref{eq:outer-loop}, can be re-written as a two step log-likelihood objective with an $l_2$ regularization term for meta-parameter $\beta$ as done in Eq. \ref{eq:outer-rewrite}.
\begin{align}
\label{eq:outer-rewrite}
\beta^* &= \arg\min_{\beta} \;  -\sum_{i=1}^{n} \C{L}_{\C{D}_i^Q}(\kappa_i^*, \lambda_i^*, \beta) + \frac{1}{2} \|\beta\|_2^2 
\end{align}

We further explain the role of $\kappa_i$, since it is the crux of our approach. For any task, it approximates the posterior distribution of the latent variables $z^x_i$ given the covariates. As can be seen from the first two terms in Eq. \ref{eq:deriv-elbo-zy}, it matches the objective for a Variational Auto-Encoders (\textsc{vae}s) \cite{kingma2013auto}. $\kappa_i$ stores the latent representation of the covariates of the task. To further simplify our objective, we ignore the third term in Eq. \ref{eq:deriv-elbo-zy}, allowing us to simply write optimal $\kappa_i^*$ as a function of only covariates of a task. As we noted in Assms.~\ref{assm:z_x}, we make $\kappa_i^*$ to consist the mean $\gamma_\mu(\C{D}_i^{S}; \beta)$ and std. deviation $\gamma_\sigma(\C{D}_i^{S}; \beta)$ of a normal distribution. Here, $\gamma_\mu(\cdot, \beta), \gamma_\sigma(\cdot, \beta)$ represent neural networks which take as input the covariates of the support set $\C{D}_i^{S}$ and output $\kappa_i^*$. It is important to note that even though the parameters of $\gamma_\mu, \gamma_\sigma$ are task-agnostic, the variational parameter $\kappa_i^*$ is still different for each task. Using this, we are able to further divide Eq.~\ref{eq:inner-loop} to optimize for $\kappa_i^*$ and $\lambda_i^*$ separately as follows
\begin{align}
\label{eq:inner-rewrite-x}
\kappa_i^* &= (\kappa_i^*(\mu), \kappa_i^*(\sigma)) = (\gamma_\mu(\C{D}_i^{S}; \beta), \gamma_\sigma(\C{D}_i^{S}; \beta))\\
\label{eq:inner-rewrite-y}
\lambda_i^* &=  \arg\min\limits_{\lambda_i} \; -\C{L}_{\C{D}_i^{S}}(\kappa_i^*, \lambda_i, \beta)
\end{align}



Having identified $\kappa_i^*$ we now describe the optimization algorithm for $\lambda_i$ in Eq.~\ref{eq:inner-rewrite-y}. Note that we ignored the $\V{y}_j^{(i)}$ while optimizing for $\kappa_i$, however, we use the latent representation of the covariates (ie $\kappa_i^*$) to inform the distribution of $z_i^y$ and hence the parameters $\lambda_i$. Following Assm.~\ref{assm:z_y}, the task-specific random variable $z^y_i$ represents the parameters of the neural network for the $i^{th}$ task, that takes as input $\V{x}_j^{(i)}$ and outputs a prediction $\V{\hat{y}}_j^{(i)}$. 
To obtain $\lambda_i^*$ it is sufficient to only minimize the objective $-\R{E}_{q(z_i^x; \kappa^*)}\C{L}_{\C{D}_i}'(z_i^x, \lambda_i, \beta)$. In Eq.~\ref{eq:deriv-elbo-zy-2}, the \textsc{kl} term acts as a regularizer in the optimization objective for $\lambda_i$. 
Since the most common algorithm for optimization is Stochastic Gradient Descent (\textsc{sgd}) many meta-learning algorithms avoid the \textsc{kl} term by choosing a regularization specific to \textsc{sgd}. 
However, in most works \cite{ravi2018amortized, finn2018probabilistic, kim2018bayesian}, the \textsc{kl} term is a function of only the meta-parameter $\theta$ (or $\beta$ given Assm.~\ref{assm:theta}). Hence the regularization is induced by letting the initialization for the optimization of $\lambda_i$ (given by $\lambda_i^{(0)}$) be determined by $\beta$.
But, in our framework, we realize that the \textsc{kl} term is a function of both $\beta$ and the latent variable $z^x_i$ (the task specific covariate distribution).
Hence we model the initialization  $\lambda_i^{(0)}$ using a neural-network whose parameters are task-agnostic and subsumed in $\beta$ and thus without loss of expressivity $\lambda_{i}^{(0)}=f_\beta(z^x_i)$ . 
Thus, the optimal parameters of the variational approximation ($\lambda_i^*$) would be given by performing $K$ steps of \textsc{sgd} on the \textsc{mle} objective in Eq.~\ref{eq:deriv-elbo-zy-2} with \textsc{kl} term replaced by the initialization $f_\beta(z^x_i)$.
\begin{align}
    \label{eq:deriv-elbo-zy-2-rewrite}
    \lambda_i^{*} = \R{E}_{q(z_i^x; \kappa_i^*)} \; & \textsc{sgd}\footnotemark(l_{\C{D}_i}(\lambda_i), \lambda_i^{(0)} = f_\beta(z^x_i), K) \\
    l_{\C{D}_i}(\lambda_i) &=  -
   \sum_{j=1}^{m'} \log p(\V{y}_j^{(i)} | \V{x}_j^{(i)}, \lambda_i) \nonumber
\end{align}

\footnotetext{$\textsc{sgd}(l(\lambda),\lambda^{(0)},K)$ is the parameter obtained by performing $K$ steps of \textsc{sgd} on the objective $l(\lambda)$ with initialization $\lambda^{(0)}$.}

The expectation in Eq~\ref{eq:deriv-elbo-zy-2-rewrite} is computed using monte-carlo approximation. We find that sampling a single value of $z_i^x \sim q(z_i^x; \kappa_i^*)$ is sufficient to optimize for $\lambda_i$.


\begin{algorithm}[H]
\caption{Meta-training Algorithm}
\begin{algorithmic}
 \STATE Given: $n$ datasets: $\{\C{D}_i^S, \C{D}_i^Q\}_{i=1}^{n}$, learning rates: $\eta_0, \eta_1$, number of update steps: $K$.  \\
 \STATE $p(\theta), p(z^x_i|\theta) \gets \C{N}(\V{0}, \V{I})$     
 \FOR{i=1 to n}
        \STATE $\kappa_i^*(\mu), \kappa_i^*(\sigma) \gets \gamma_\mu(\C{D}_i^{S};\beta), \gamma_\sigma(\C{D}_i^{S};\beta)$ \\
        \STATE $\lambda_i^{(0)} = f_\beta(\kappa_i^*(\mu) + \epsilon \circ \kappa_i^*(\sigma))$; $\;\;\;\epsilon \sim  \C{N}(\V{0}, \V{I})$   \\
        \FOR{t=0 to K-1}
            \STATE $\lambda_i^{(t+1)} \gets  \lambda_i^{(t)} - \eta_0 \nabla_{\lambda_i^{(t)}} \sum\limits_{j=1}^{m'} \log p(\V{y}_j^{(i)} | \lambda_i^{(t)} , \V{x}_j^{(i)})  $ \\
        \ENDFOR
        \STATE $\lambda_i^*  \gets  \lambda_i^{(K)}$ 
        

        
        
        
         \STATE $\beta \gets  \beta - \eta_1 \nabla_{\beta} \left[\textsc{obj}(\beta)\right]$
        
 \ENDFOR
 \end{algorithmic}
 \label{alg:train}
\end{algorithm}

Finally, we note that the meta-parameter $\beta$ constitutes the parameters of the network $f_\beta$ which determines the initialization $\lambda_i^{(0)}$ as well as the parameters of $\gamma_\mu(\cdot; \cdot), \gamma_\sigma(\cdot; \cdot)$ which output $\kappa_i^*$.  
Thus, $\beta$ is optimized to jointly maximize the likelihood of the covariates of a sequence of tasks as well as for learning to choose covariate dependent initializations suitable for few-shot adaptation.
For the optimization objective in Eq.~\ref{eq:outer-rewrite}, we use the standard re-parameterization trick ($2^{nd}$ step of the outer \textit{for-loop} in Algorithm~\ref{alg:train}) commonly used in \textsc{vae}s. This is done so as to be able to differentiate through the expectation over $q(z_i^x; \kappa_i^*)$ in Eqs.~\ref{eq:deriv-elbo-zy}, ~\ref{eq:deriv-elbo-zy-2-rewrite}. Finally, continuing from Eq.~\ref{eq:outer-rewrite}, we define below the final meta-learning objective in terms of $\beta$ that we optimize using gradient descent in the outer loop. Using re-parameterization, we have $z^x_i = \gamma_\mu(\C{D}_i^{Q};\beta) + \epsilon \circ \gamma_\sigma(\C{D}_i^{Q};\beta) \; \epsilon \sim  \C{N}(\V{0}, \V{I})$. In the objective, $\C{L}^{(i)}_{R}$ stands for reconstruction loss of the covariates and come from first term in Eq.~\ref{eq:deriv-elbo-zy}, the $\C{L}^{(i)}_{\textsc{kl}}$ is the second term in the same equation representing \textsc{kl} divergence between two normal distributions and finally $\C{L}^{(i)}_{SGD}$ represent the likelihood of query set as in Eq.~\ref{eq:deriv-elbo-zy-2} using the adapted parameters $\lambda_i^*$ of Eq.~\ref{eq:deriv-elbo-zy-2-rewrite}
\begin{align}
    \label{eq:final-obj}
    & \textsc{obj}(\beta) = \alpha_{l_2}\|\beta\|_2^2  -\alpha_{R} \sum_{i=1}^{n}  \overbrace{\sum_{j=m'}^{m} \log p(\V{x}_j^{(i)} | z^x_i)}^{\C{L}^{(i)}_{R}} \\ \nonumber
    & + \alpha_{KL} \sum_{i=1}^{n} \underbrace{\textsc{kl}(\C{N}(\gamma_\mu(\C{D}_i^{Q};\beta),\gamma_\sigma(\C{D}_i^{Q};\beta)) || \C{N}(\V{0}, \V{I}))}_{\C{L}^{(i)}_{KL}}  \\ \nonumber
    & - \sum_{i=1}^{n}  \underbrace{\sum_{j=m'}^{m} \log p(\V{y}_j^{(i)} | \V{x}_j^{(i)}, \lambda_i^*)}_{\C{L}^{(i)}_{SGD}}
\end{align}

Although, based on current assumptions, the coefficients $\alpha_{l_2}, \alpha_{R}, \alpha_{\textsc{kl}}$ should be 1 theoretically. However, in practice we choose their values based on validation set. Note that this does not violate the theory, as it corresponds to different priors ($p(\theta), p(z_i^x|\theta)$) in Assms.~\ref{assm:theta}, ~\ref{assm:z_x} which were taken as standard normal in derivations for simplicity.
The step-by-step procedure for the meta-training and meta-testing phases are given by Algorithm~\ref{alg:train} and Algorithm~\ref{alg:test} respectively. 

\begin{algorithm}[H]
\caption{Meta-testing Algorithm on test task $\C{T}$}
\begin{algorithmic}
 \STATE Given: dataset $\C{D}_{\C{T}} = \{\V{x}_j^{(\C{T})}, \V{y}_j^{(\C{T})}\}_{j=1}^{m}$, parameter: $\beta^*$, learning rate $\eta_0$, number of update steps: K.
 \STATE $\kappa_{\C{T}}^*(\mu), \kappa_{\C{T}}^*(\sigma) \gets \gamma_\mu(\C{D}_{\C{T}};\beta^*), \gamma_\sigma(\C{D}_{\C{T}};\beta^*)$ \\
        \STATE $\lambda_{\C{T}}^{(0)} = f_{\beta^*}(\kappa_{\C{T}}^*(\mu) + \epsilon \circ \kappa_{\C{T}}^*(\sigma))$; $\;\;\;\epsilon \sim  \C{N}(\V{0}, \V{I})$   \\
        \FOR{t=0 to K-1}
            \STATE $\lambda_{\C{T}}^{(t+1)} \gets  \lambda_{\C{T}}^{(t)} - \eta_0 \nabla_{\lambda_{\C{T}}^{(t)}} \sum\limits_{j=1}^{m} \log p(\V{y}_j^{(\C{T})} | \lambda_{\C{T}}^{(t)} , \V{x}_j^{(\C{T})})  $ \\
        \ENDFOR
        \STATE $\lambda_{\C{T}}^*  \gets  \lambda_{\C{T}}^{(K)}$ 
 \end{algorithmic}
 \label{alg:test}
\end{algorithm}

\section{Experiments and Results}
\label{sec:experiments}

In order to first litmus test our approach on a simpler task we begin by evaluating it on a synthetic regression dataset which is a modified version of the one proposed by \citet{vuorio2019multimodal} and follow it up with experiments on four few-shot image recognition benchmarks. We shall now describe our baselines followed by a discussion on the experimental setup for our regression and classification experiments. 

\paragraph{Baselines} Since our algorithm revolves around an initialization based approach (as opposed to learning a shared backbone as in \citet{lee2019meta,bertinetto2018metalearning}), we limit our comparisons to: the gradient based meta-learning approach \textsc{maml} introduced by \citet{finn2017model} and its first-order approximation \textsc{reptile} \cite{nichol2018on}. Additionally, we also chose as baselines: Amortized \textsc{maml} \cite{ravi2018amortized} and \textsc{m-maml} \cite{vuorio2019multimodal} which are exemplars of the Bayesian and task-specific initialization based approaches respectively. These methods either model the task-parameters as Bayesian random variables (former) or adapt the parameters of the optimizer based on the input dataset (latter) and hence warrant a close comparison with our approach.

\subsection{Regression}

Most meta-learning algorithms have been tested on regression datasets where the covariate distribution is same across all tasks \cite{vuorio2019multimodal}. In this setting, we instead describe how we suitably modify the original dataset so that there exists a structure over the set of covariate distributions across tasks. This enables us to fairly evaluate our method against other baselines in this potentially more challenging setting.

\begin{table}[!ht]
    \centering
    \begin{tabular}{c c c c}
    \textbf{Model}     & \textit{sine}  & \textit{sine-quad-linear} & \textit{five} \\\toprule
    \textsc{maml} & 0.05 & 1.27  & 1.69   \\
    Amortized \textsc{maml} & 0.07 & 1.39  & 1.13   \\
    \textsc{m-maml} &  0.04 & 0.59  & 0.93\\
    Ours  & \textbf{0.008} & \textbf{0.39} & \textbf{0.89}\\
    \end{tabular}
    \vspace{0.5em}
    \caption{Comparison of post adaptation test performance (\textsc{mse} loss) on three regression datasets when the true hypothesis class depends on the covariate distribution.}
  \label{tab:results-1}
\end{table}

\begin{table}[!ht]
    \centering
    \begin{tabular}{c c c c}
    \textbf{Model}     & \textit{sine}  & \textit{sine-quad-linear} & \textit{five} \\\toprule
    \textsc{maml} & 0.04 & 1.15  & 1.73   \\
    Amortized \textsc{maml} & 0.07 & 1.41  & 1.08   \\
    \textsc{m-maml} &  \textbf{0.03} & 0.51  & 0.88\\
    Ours  & 0.04 & \textbf{0.51} & \textbf{0.84}\\
    \end{tabular}
    \vspace{0.5em}
    \caption{Comparison of post adaptation test performance (\textsc{mse} loss) on three regression datasets when the true hypothesis class is independent of the covariate distribution.}
    \label{tab:results-2}
\end{table}

\begin{table*}[!ht]
    \centering
    {\setlength{\tabcolsep}{0.13em}
    \begin{tabular}{ c  c  c  c  c  c  c}
    \textbf{Model}     & \multicolumn{2}{c}{\textbf{miniImageNet}}  & \multicolumn{2}{c}{\textbf{CUB}} & \multicolumn{2}{c}{\textbf{CIFAR-FS}}  \\\toprule
    &\textbf{5w5s} & \textbf{5w1s} & \textbf{5w5s} & \textbf{5w1s}  & \textbf{5w5s} & \textbf{5w1s} \\ \midrule
    \textsc{maml}*           &  $63.11 \pm 0.42 \%$ &   $48.70 \pm 1.84 \%$ &  $72.09 \pm 0.76 \%$ &   $55.92 \pm 0.95 \%$ &  $71.50 \pm 1.00 \%$ & $58.90 \pm 1.90 \%$  \\
    \textsc{Reptile}*        &  $62.74 \pm 0.37 \%$ &   $47.07 \pm 0.26 \%$ &  $74.17 \pm 0.58 \%$ &   $53.79 \pm 0.59 \%$ &  $72.38 \pm 1.04 \%$ & $60.02 \pm 1.21 \%$ \\
    Amortized \textsc{maml} &  $63.15 \pm 0.42 \%$ &   $47.59 \pm 0.47 \%$ &  $73.85 \pm 0.80 \%$ &   $56.23 \pm 1.12 \%$ &  $71.73 \pm 0.95 \%$ & $59.70 \pm 0.79 \%$ \\
    \textsc{m-maml}          &  $62.81 \pm 0.59 \%$ &   $49.09 \pm 0.52 \%$ &  $74.58 \pm 1.01 \%$ &   $57.28 \pm 1.08 \%$ &  $\mathbf{73.01 \pm 0.67 \%}$ & $\mathbf{60.93 \pm 0.84 }\%$ \\ \midrule
    Ours                    &  $\mathbf{64.81 \pm 0.59 \%}$ &   $\mathbf{49.89 \pm 0.40 \%}$ &  $\mathbf{75.13 \pm 0.97 \%}$ &   $\mathbf{57.99 \pm 0.88 \%}$ &  $72.03 \pm 0.74 \%$ & $60.17 \pm 0.76 \%$ \\
    \end{tabular}}
    \caption{Comparison to prior work on \textbf{miniImageNet}, \textbf{CUB} and \textbf{CIFAR-FS} datasets denoting the mean few-shot classification accuracies (in $\%$) along with their $95\%$ confidence intervals on \textbf{5w5s} and \textbf{5w1s} tasks (best model chosen based on validation performance). For algorithms marked with ``*", evaluations on reported datasets have been borrowed from original works and the results on the rest are based on our re-implementation.}
  \label{tab:classification-results}
\end{table*}

\paragraph{Datasets} The covariate distribution for each task $\C{T}$ is given by a normal $\C{N}(\mu_\C{T}, \sigma_{\C{T}})$ whose parameters are sampled from a discrete distribution over $P$ pairs $\{(\mu_p, \sigma_p)\}_{p=1}^{P}$.
The pairs $p_1, \dots p_P$ are fixed at the beginning once they are sampled from a pair of independent uniform priors, $p_i \sim (\C{U}(-10, 10), \; \C{U}(0, 10))$. 
The parameters of the discrete distribution are sampled from a Dirichlet prior.  Following \cite{vuorio2019multimodal}, the optimal hypothesis for each task is sampled from one of the five modalities (or hypothesis classes): \textit{sine}, \textit{linear}, \textit{quad}, \textit{transformed-L$_1$} and \textit{tanh}. For each task, having chosen a hypothesis class, the parameters of the optimal hypothesis (like slope of a linear function) is determined based on samples from uniform distributions (more details in the Appendix). 



\paragraph{Discussion} We consider two cases, \textit{first} where there exists a relation between the parameters (mean, variance) of the covariate distribution and the optimal hypothesis class chosen for a task and \textit{second} when the optimal hypothesis class is chosen independent of the mean, variance of the covariate distribution. 
\textbf{Case-I (specific relation):} This setting conforms to the case when $z_y {\not\!\perp\!\!\!\perp} z_x | \theta$ in Fig.~\ref{fig:gm_CPMAML}. We experiment with three different meta-distributions (\textit{sine}, \textit{sine-quad-linear}, \textit{five})\footnote{Details of each can be found in the Appendix.} which are of different complexities owing to the number of modes in it.  Table~\ref{tab:results-1} highlights the Mean Squared Errors (\textsc{mse}) achieved on these datasets from which we can infer the following: when there exists a relation between the true hypothesis class and the covariate distribution of the task, our approach performs significantly better than other state-of-the-art approaches for regression. 
The Bayesian model Amortized \textsc{maml} performs poorly since unlike our approach it fails to model the latent relationship between the covariate distribution and the posterior over the modes in the meta-distribution. 
\textbf{Case-II (Independent):} This setting conforms to the case when $z_y \perp\!\!\!\perp z_x | \theta$ in Fig.~\ref{fig:gm_CPMAML}. In Table~\ref{tab:results-2} we demonstrate that the performance of our approach is no worse (if not better) than other methods which assume the independence by default.

\begin{table}[!ht]
    \centering
    {\setlength{\tabcolsep}{0.4em}
    \begin{tabular}{ c  c  c}
    \textbf{Model}     & \textbf{5w5s}  & \textbf{5w1s} \\\toprule
    \textsc{maml}           &  $51.70 \pm 0.90 \%$ &   $39.90 \pm 1.80 \%$  \\
    \textsc{Reptile}        &  $52.38 \pm 0.51 \%$ &   $41.95 \pm 1.33 \%$  \\
    Amortized \textsc{maml} &  $50.40 \pm 1.11 \%$ &   $39.08 \pm 1.76 \%$  \\
    \textsc{m-maml}          &  $52.23 \pm 1.03 \%$ &   $41.90 \pm 1.42 \%$  \\ \midrule
    Ours $(\alpha_{R}, \alpha_{KL})$ & & \\\midrule
    $0.2,10^{-2}$   &  $49.99 \pm 1.25 \%$ &  $39.65 \pm 0.40 \%$ \\
     $10^{-2},10^{-3}$   &   $52.10 \pm 0.98 \%$ &  $41.81 \pm 0.32 \%$ \\
    \end{tabular}}
    \caption{Comparison to prior work on \textbf{FC-100} dataset denoting the mean few-shot classification accuracies (in $\%$) along with their $95\%$ confidence intervals on \textbf{5w5s} and \textbf{5w1s} tasks.}
  \label{tab:classification-results-FC100}
\end{table}


\subsection{Classification Results}
\label{subsec:classification-results}

We demonstrate the benefits of modeling the covariate distribution for few-shot recognition by highlighting the suitability of the auxiliary objective which accounts for the log-likelihood over the support samples in a task. As seen in Table \ref{tab:classification-results}, optimizing for this additional objective in turn influences the choice of the initialization in a way that leads to a more generalizable solution.   

\paragraph{Datasets} We experiment on four few-shot recognition benchmarks: \textbf{miniImageNet} \cite{vinyals2016matching}, \textbf{CUB} \cite{WahCUB_200_2011}, \textbf{CIFAR-FS} \cite{bertinetto2018metalearning} and \textbf{FC-100} \cite{oreshkin2018tadam}. The most popular among these is \textbf{miniImageNet}, which has a split of $64$ train, $16$ validation and $20$ test classes, each with $600$ examples. \textbf{CIFAR-FS} and \textbf{FC-100} have both been derived from the CIFAR-100 \cite{krizhevsky2009learning} dataset where each class has $600$ images of size $32\times32$. \textbf{FC-100} is a harder benchmark since the construction of it involves dividing the original $100$  classes into $20$ super-classes with $12,4$ and $4$ super-classes in the train, validation and test splits respectively. The severely reduced information overlap between the train and test distributions raises the difficulty of few-shot adaptation. Each task sampled from the meta-distribution comprises of \textit{n-shot} support samples for each of the \textit{n-way} classes. Additionally, \textit{n-query} points are also provided for each class, to evaluate the output of the meta-learner on the input task. In our experiments, we evaluate algorithms on two types of tasks, (i) \textbf{5w5s}: \textit{n-way} $=5$, \textit{n-shot} $=5$, \textit{n-query} $=15$ and (ii) \textbf{5w1s}: \textit{n-way} $=5$, \textit{n-shot} $=1$, \textit{n-query} $=15$. During training, we match the number of classes for each task (\textit{n-way}) and the size of the support and query sets in the meta-train and meta-test phases. 

\paragraph{Discussion} From Table \ref{tab:classification-results} we clearly note improvements $>1.5\%$ on  \textbf{5w5s}  and $\approx 0.8\%$ on \textbf{5w1s} tasks sampled from  \textbf{miniImageNet}. Similar gains can also be noted for the \textbf{CUB} dataset where the model needs to learn localized features for a more fine-grained classification between the various species of birds. Our loss objective includes the reconstruction and \textsc{kl} terms for the covariates which is one of the major differences compared to \textsc{m-maml}. Since we observe gains over \textsc{m-maml}, we can conclude that the performance boost can be attributed exclusively to the log-likelihood objective over the task covariates (as opposed to other factors like choice of architecture). 

Owing to the super-class level train/test splits, \textbf{FC-100} suffers from significantly disjoint covariate distributions for the meta-train and meta-test tasks. Our hypothesis is that the objective we propose can specifically lead to poorer generalization (over traditional \textsc{maml}) in this case. Since per-task initializations are inferred directly from the covariate distribution, the network may be initialized in a completely different region for the meta-test tasks. This is less of a concern in a dataset like \textbf{miniImageNet} or \textbf{CUB} where the parametric mapping from the covariates to the network initialization happens to generalize across the splits. Moreover, from Table \ref{tab:classification-results-FC100} we note that when $\alpha_R, \alpha_{KL} \rightarrow 0$, we achieve the performance of \textsc{m-maml}.

\paragraph{Implementation} Our final loss formulation involves optimizing the four term objective in Eq. \ref{eq:final-obj}. In this, the parameters of the conditional $p(y|x)$ is modeled using a $100$ dimensional, $3-$layer neural network for regression and a Conv64 backbone \cite{finn2017meta, nichol2018first} for classification. The networks $\gamma_\mu(\cdot; \beta), \gamma_\sigma(\cdot; \beta)$ which take as input the sequence of covariates in a dataset $\C{D}_i^S$ are modeled using an \textsc{rnn} with a hidden size of $28$ and $256$ for regression and classification respectively. 
The mapping of the \textsc{rnn} output to $\lambda_i^{(0)}$, given by $f_\beta(\cdot)$, is modeled using a neural-network and is modulated in the same manned as \textsc{m-maml},
which gives us the final initialization of the network parameters. 
We didn't extensively fine-tune for the $\alpha_*$ weights in Eq. \ref{eq:final-obj} but the following values worked considerably well for all classification datasets: $\alpha_R=0.2, \alpha_{KL}=0.01, \alpha_{l_2}=0.0005$. For regression, $\alpha_{KL}=0.1$ was used instead.


\section{Conclusion}
\label{sec:conclusion}

Cognizant of the fact that the generalization performance of few-shot algorithms depends on a varying number of factors ranging from sample size, hypothesis class complexity to the optimization algorithm, shift in train/test meta-distributions; in this work we focus our efforts on improving meta-learning algorithms by using the covariates to infer the adapted parameters via a principled Bayesian approach. We begin by deriving \textsc{elbo} bounds for the hierarchical Bayes formulation and follow it up with a meta-learning algorithm to infer the posterior over the network parameters. Finally, we demonstrate the performance of our approach on a synthetic regression dataset in addition to four standard classification benchmarks. In future, we plan to extend our work to deeper backbones like Resnet-12 and also to amortized inference based meta-learners \cite{lee2019meta, bertinetto2018meta}.



\bibliography{main.bib}
\bibliographystyle{icml2020}

\end{document}